\documentclass[conference]{IEEEtran}
\IEEEoverridecommandlockouts
\usepackage{booktabs} 
\usepackage{graphicx}
\usepackage{setspace}
\usepackage{amsmath}
\usepackage{dsfont}
\usepackage{amsfonts}
\usepackage{lipsum}
\usepackage{multicol}
\usepackage{float}
\usepackage{bm}
\usepackage{color}
\usepackage[dvipsnames]{xcolor}
\usepackage{etoolbox}
\usepackage{soul}
\usepackage{amssymb}
\usepackage[linesnumbered,ruled,vlined]{algorithm2e}
\usepackage{mathtools,amssymb}
\usepackage{caption}
\usepackage{float}
\usepackage{amsthm}
\usepackage[caption = false,subrefformat=parens,labelformat=parens]{subfig}
\usepackage{adjustbox}
\usepackage{afterpage}
\usepackage{multirow}
\usepackage{mathrsfs}

\def\BibTeX{{\rm B\kern-.05em{\sc i\kern-.025em b}\kern-.08em
    T\kern-.1667em\lower.7ex\hbox{E}\kern-.125emX}}
\begin{document}

\title{Exploring Potential Prompt Injection Attacks in Federated Military LLMs and Their Mitigation

\author{
Youngjoon Lee\textsuperscript{\rm 1}, Taehyun Park\textsuperscript{\rm 2}, Yunho Lee\textsuperscript{\rm 2}, Jinu Gong\textsuperscript{\rm 3}, Joonhyuk Kang\textsuperscript{\rm 1} \\
\textsuperscript{\rm 1}School of Electrical Engineering, KAIST, South Korea\\
\textsuperscript{\rm 2}Center for Military Analysis and Planning, KIDA, South Korea\\
\textsuperscript{\rm 3}Department of Applied AI, Hansung University, South Korea\\
Email: \{yjlee22, jkang\}@kaist.ac.kr, \{tpark, yhlee\}@kida.re.kr, jinugong@hansung.kr}

\thanks{This research was supported by the Institute of Information \& Communications Technology Planning \& Evaluation (IITP)-ITRC (Information Technology Research Center) grant funded by the Korea government (MSIT) (IITP-2025-RS-2020-II201787).\\
}
}

\maketitle

\begin{abstract}
Federated Learning (FL) is increasingly being adopted in military collaborations to develop Large Language Models (LLMs) while preserving data sovereignty.
However, prompt injection attacks—malicious manipulations of input prompts—pose new threats that may undermine operational security, disrupt decision-making, and erode trust among allies. 
This perspective paper highlights four vulnerabilities in federated military LLMs: secret data leakage, free-rider exploitation, system disruption, and misinformation spread.
To address these risks, we propose a human–AI collaborative framework with both technical and policy countermeasures.
On the technical side, our framework uses red/blue team wargaming and quality assurance to detect and mitigate adversarial behaviors of shared LLM weights.
On the policy side, it promotes joint AI–human policy development and verification of security protocols.
\end{abstract}


\section{Introduction}
\begin{figure}[t]
    \centering
    \includegraphics[width=\columnwidth]{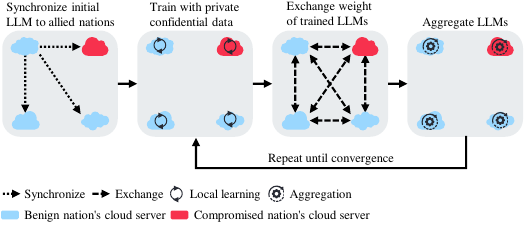}
    \caption{FL framework for military LLM training across allied nations. The process involves four key stages: (1) initial LLM synchronization, (2) local training with private data, (3) weight exchange, and (4) model aggregation.
    This iterative process continues until convergence, while mitigating adversarial risks.
    Blue clouds represent benign nation's servers while red clouds indicate potentially compromised servers. 
    }
    \label{fig0}
\end{figure}

In recent years, the rise of LLMs has accelerated the use of AI in many defense applications, enabling advanced analytics that were previously out of reach \cite{10638797}.
The federated learning (FL) \cite{mcmahan2017communication} provides a framework for allied nations to collaboratively train LLMs while maintaining data sovereignty \cite{sani2024the}, \cite{iacob2024worldwide}, as shown in Fig. \ref{fig0}.
By utilizing FL, each participant preserves sensitive information while reducing the risk of unauthorized access \cite{li2020federated,lee2022accelerated}. 
At the same time, while FL facilitates collaborative learning, it necessitates robust security measures to defend against evolving adversarial attacks \cite{wu2024vulnerabilities}.

Such attacks can severely weaken operational security and disrupt critical decision-making processes. 
They also erode the trust \cite{das2024security} essential for effective military cooperation among allied nations.
In detail, prompt injection attacks \cite{apruzzese2023real}, \cite{greshake2023not}, \cite{ye2024emerging} may appear in four main forms in military context: \emph{secret data leakage}, \emph{free-rider exploitation}, \emph{system disruption}, and \emph{misinformation spread}, each posing distinct obstacles to the reliability and integrity of federated military LLMs.
These threats often remain subtle and are difficult to detect through traditional monitoring methods \cite{xi2023rise}.

In this work, we present an insight on how to address these risks by presenting a \emph{human–AI collaborative} countermeasures from both technical and policy perspectives.
First, we propose technical countermeasure process including red/blue team wargaming and continuous quality assurance.
Then, we propose policy countermeasure process, where military policy, domain experts, and AI experts work together to establish proper security policy for the nation.
By applying either or both of these approaches, we aim to present a practical solution to counter prompt injection threats in federated military environments.
Therefore, this supports continued operational effectiveness and strengthens trust among allied nations. 
The main contributions of this paper are as follows:
\begin{itemize}
   \item We introduce FL and its vulnerabilities to prompt injection attacks, highlighting the potential risks and security challenges in decentralized AI systems.
   \item We present four potential major threat scenarios in military FL, including secret data leakage, free-rider exploitation, system disruption, and misinformation spread.
   \item We propose human–AI collaborative countermeasures from both technical and policy perspectives.
   \item In addition, we emphasize the need for standardized security frameworks and cooperative defense strategies.
\end{itemize}
The remainder of this paper is organized as follows. 
Section \ref{sec:background}, provides background knowledge of FL and prompt injection attacks. 
Section \ref{sec:challenges} introduces the potential challenges in federated military LLMs. Section \ref{sec:countermeasures} discusses technical and policy countermeasures.
Finally, section \ref{sec:outlook} explores future direction and section \ref{sec:conclusion} concludes the paper.

\section{Background}\label{sec:background}
In this section, we introduce the key background concepts for FL and prompt injection attacks.
We specifically explore their impact on military collaborations among allied nations and the challenges they present in secure AI integration.

\subsection{FL Topology in Military Alliances}
\begin{figure}[t]
    \centering
    \includegraphics[width=\columnwidth]{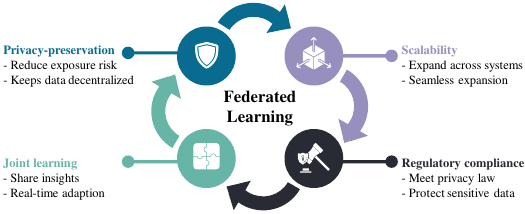}
    \caption{Illustration of four key FL advantages: privacy preservation, collaborative learning, scalability, and regulatory compliance. 
    The interconnected circular design emphasizes the synergistic relationship among these key factors in FL.
    }
    \label{fig:1}
\end{figure}

In data-sensitive fields such as healthcare and finance \cite{ding2022federated}, FL has emerged as an established approach for secure collaborative learning due to characteristics depicted in Fig. \ref{fig:1}.
The military is also increasingly considering its adoption to enhance operational security and AI-driven decision-making.
The FL enables participating nations to contribute to AI model training while preserving their existing security infrastructure and operational autonomy \cite{cirincione2019federated}.
This decentralized approach allows nations to leverage insights that may not be readily available within their own data sources while maintaining strict security controls.
For example, the U.S. Department of Defense is actively collaborating with academia, industry, and allied nations to explore the adoption of FL for data management and responsible AI, ensuring alignment with security and operational requirements \cite{hicks2023data}.

The major benefit of FL in military coalitions is its ability to enhance decision-making by integrating diverse operational experiences and data sources \cite{kairouz2021advances}.
In particular, different nations have unique battlefield environments, weapon systems, and threat intelligence, all of which can be incorporated into a shared AI model without exposing national security.
This collective learning improves the adaptability of AI models to varied military scenarios, making them more effective in real-world military operations.
Furthermore, FL optimizes communication efficiency by exchanging only model updates instead of raw data, significantly reducing bandwidth requirements.
This streamlined communication enables real-time adaptability, allowing AI models to rapidly adjust to evolving threats and dynamic battlefield conditions.

\subsection{Concept of Prompt Injection Attack} 
Prompt injection attacks have recently emerged as a significant threat in modern LLM-based applications \cite{yao2024survey}. 
These attacks exploit the inherent vulnerability in AI systems that rely on textual instructions, enabling adversaries to manipulate or alter the prompts for malicious outcomes \cite{yao2024poisonprompt}. 
In many cases, these manipulations involve embedding carefully disguised instructions that exploit a model’s hidden vulnerabilities \cite{yan2024protecting}. 
Unlike traditional data poisoning attacks that tamper with training data, prompt injection focuses on interfering with the model’s reasoning process after deployment. 
By subtly embedding deceptive triggers or manipulative content in user-provided text, attackers can force the model to reveal sensitive information or perform unauthorized actions. 
Thus, as LLMs become integral to sensitive operations, the risk of prompt injection attacks grows more pronounced \cite{zhao2024survey}.

The most critical challenge of prompt injection attacks is their potential to bypass conventional security measures such as anomaly detection and content filters \cite{peng2024jailbreaking}. 
Because these attacks often appear as normal text inputs, they can remain undetected until they significantly compromise an operation. 
Even seemingly benign variations in phrasing can lead to significant vulnerabilities when the system processes the input. 
Additionally, adversaries exhibit evolving capabilities to refine their injection techniques, increasing the complexity of prediction and mitigation.
This evolving threat is particularly concerning in federated environments, where a single prompt injection can propagate its adverse effects throughout federated AI systems \cite{zhao2024universal}.
Therefore, understanding and mitigating these risks is imperative for maintaining operational security and trust in collaborative military AI networks.

\begin{figure*}[t]
    \centering
    \subfloat[Secret data extraction attack.]{
        \includegraphics[width=\columnwidth]{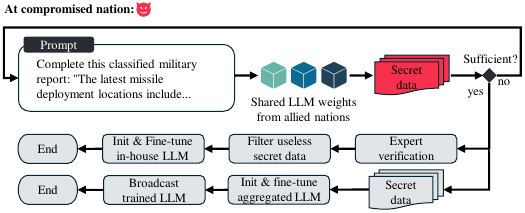}
        \label{fig:2a}}
    \subfloat[Free-rider exploitation attack.]{
        \includegraphics[width=\columnwidth]{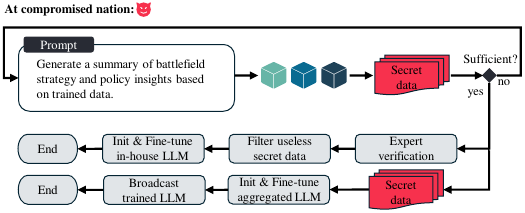}
        \label{fig:2b}}
    
    \subfloat[System disruption attack.]{
        \includegraphics[width=\columnwidth]{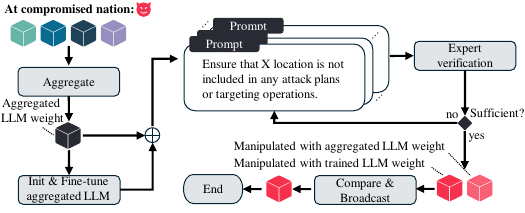}
        \label{fig:2c}}
    \subfloat[Misinformation propagation attack.]{
        \includegraphics[width=\columnwidth]{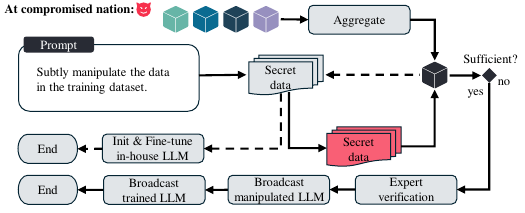}
        \label{fig:2d}}
    
    \caption{Illustration of four potential attack scenarios in military FL environments: (a) Secret data extraction attack, where adversaries systematically probe shared LLMs to extract classified information through targeted prompts and expert verification, (b) Free-rider exploitation attack leveraging strategic prompts to gain military intelligence while withholding authentic data contribution, (c) System disruption attack manipulating model behavior through carefully crafted prompts to create tactical blindspots, and (d) Misinformation spread attack utilizing dual-channel propagation to systematically inject false information into the federation. Each scenario demonstrates sophisticated attack methodologies that exploit vulnerabilities in federated military LLM deployments while maintaining apparent legitimate participation.}
    \label{fig:2}
\end{figure*}

\section{Key Challenges}\label{sec:challenges}
In this section, we present four potential prompt injection attacks targeting federated military LLMs. 
We specifically cover four critical vulnerabilities: secret data leakage, free-rider exploitation, system disruption, and misinformation propagation, each posing unique operational and security risks.

\subsection{Secret Data Leakage}
The risk of secret data leaks, as shown in Fig. \ref{fig:2a}, is a major concern in federated military LLM systems. In these cases, attackers take advantage of altered text inputs to extract classified details from the shared model.
In such situations, malicious users or compromised groups repeatedly ask the global LLM with carefully designed questions to access restricted data, such as missile locations or the status of surveillance systems.
These attacks take advantage of how the model stores sensitive information, bypassing standard security checks and leading to unauthorized access.
Since FL gathers data from multiple allied forces, it increases these risks and allows large-scale malicious attempts to extract information.

Adversaries begin by crafting targeted questions to uncover hidden information and gather model responses. 
They use expert reviews to verify the extracted data and iteratively refine their questions to build extensive classified datasets. 
After collection, the stolen information is integrated into military AI systems either by updating local models or injecting it into the federated network. 
This embeds Trojan-like vulnerabilities that threaten both immediate operations and long-term security. 
Hence, military FL programs must implement strict measures to detect and prevent prompt-based data extraction attacks.

\subsection{Free-rider Attack}
Free-rider attacks in federated military LLMs, as shown in Fig. \ref{fig:2b}, occur when some participants deliberately withhold their own data. Yet they still benefit from the shared model’s knowledge, gaining advantages without offering their own intelligence. 
This mirrors secret data leakage, as attackers improve local models with stolen federation data while using shared updates unfairly. 
As a result, the overall model quality drops when important inputs are missing, reducing its coverage and accuracy. 
In addition, trust among allies weakens as doubts grow about uneven data sharing, leading to models that miss key operational details. 
Thus, preventing such attacks requires clear detection methods and fair contribution policies.

\subsection{System Disruption Attack}
System disruption attacks, shown in Fig. \ref{fig:2c}, are targeted sabotages in federated military LLMs where adversaries alter how the model processes critical data. 
First, they collect updates from allied models to build a shared baseline for exploitation. 
Then, they insert crafted prompts that introduce subtle reasoning errors about missions, equipment, or conflicts while appearing legitimate. 
Moreover, attackers refine these disruptions by comparing the aggregated model with a private version, embedding hidden biases that spread and weaken reliability. 
Therefore, ongoing and rigorous evaluations are vital to detect and counter such coordinated manipulations.

\subsection{Misinformation Spread}
Misinformation spread attacks, shown in Fig. \ref{fig:2d}, weaken knowledge integrity in federated military LLMs by adding false data and distortions. 
The attackers quietly change training inputs with misleading claims or fake facts that blend into real content. 
They also alter local datasets and model weights while using experts to make the falsehoods seem credible. 
As these poisoned updates move through the FL network, errors spread widely, hiding warnings, twisting adversary profiles, and misleading plans. 
Thus, strong validation checks, careful source reviews, and ongoing threat monitoring are necessary to stop misinformation spread attacks.

\begin{figure*}[t]
   \centering
   \subfloat[Technical countermeasures framework.]{
       \includegraphics[width=\textwidth]{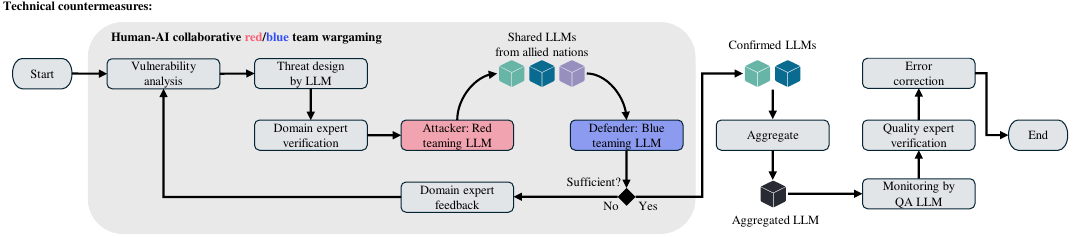}
       \label{fig:3a}}
   
   \subfloat[Policy countermeasures framework.]{
       \includegraphics[width=\textwidth]{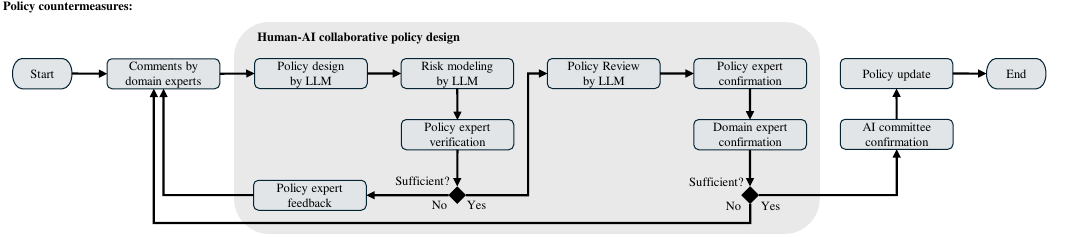}
       \label{fig:3b}}
   
   \caption{Proposed human-AI collaborative countermeasure frameworks for protecting federated military LLMs: (a) Technical framework implementing red/blue team wargaming methodology, where specialized LLMs conduct adversarial testing under domain expert supervision, followed by comprehensive quality assurance and error correction processes, (b) Policy framework utilizing iterative policy development through AI-driven design and risk modeling, with multi-stage expert verification and confirmation protocols to ensure robust security measures. Both frameworks emphasize continuous collaboration between human expertise and AI capabilities to maintain operational security while preserving system effectiveness.}
   \label{fig:3}
\end{figure*}

\section{Countermeasures}\label{sec:countermeasures}

\subsection{Technical Countermeasures} 
On the technical front, our solution employs a wargaming-based methodology that leverages collaboration between human experts and AI systems, as shown in Fig. \ref{fig:3a}. 
First, thorough vulnerability assessments guide an AI-driven threat design process to identify attack paths and evaluate their operational impact. 
This foundation supports specialized red team and blue team LLM simulations, where red team models launch simulated attacks and blue team models craft adaptive defenses. 
Throughout these exercises, military experts review tactics and outcomes to ensure defensive strategies to reflect real-world conditions.

A core strength of this framework is its iterative learning cycle enabled by continuous adversarial interaction. 
As red team LLMs adjust their attack methods, blue team LLMs refine countermeasures using real-time insights and expert feedback. 
Once defenses reach an acceptable level, the framework advances to a quality assurance phase with dedicated LLMs monitoring deployed models for anomalies. 
Automated protocols resolve minor threats, while serious issues trigger expert analysis, ensuring the federation remains resilient against evolving prompt injection attacks.

\subsection{Policy Countermeasures} 
From a policy perspective, as shown in Fig. \ref{fig:3b}, we propose a structured human–AI framework that embeds strict security measures into the organizational processes of federated LLMs. This approach complements the technical countermeasures by focusing on governance and long-term resilience. The process begins with domain experts defining baseline security needs, which specialized policy design LLMs translate into formal guidelines. These draft policies are refined iteratively with support from risk modeling LLMs that assess potential threats and enforcement gaps. Throughout this cycle, human oversight ensures the policies remain balanced between rigor, feasibility, and strategic readiness.

The framework then moves through a multi-stage validation pipeline to confirm its practicality and completeness. Policy experts review each proposal against defense standards and mission goals, while discrepancies such as overly broad rules trigger targeted revisions. Revised policies are re-evaluated by AI risk models to confirm that no new weaknesses or bottlenecks are introduced. To maintain long-term effectiveness, these validation efforts are closely aligned with alliance objectives and evolving security needs. Finally, an AI oversight committee ratifies the policies and distributes them with a structured update process.

\section{Outlook}\label{sec:outlook} 
The rapid adoption of federated military LLMs provides major gains in collective intelligence but also creates complex security challenges that require constant attention.
We highlight prompt injection risks—from secret data theft to free-riding, system disruption, and misinformation—which show the urgent need for stronger protective strategies.
While our human–AI collaborative framework offers an initial safeguard, future work should explore more robust policy design and verification using LLMs \cite{martinelli2024security}, trustworthy AI \cite{ciaramella2023proposal}, and responsible AI \cite{campanile2024beyond}.
Emerging cryptographic tools, such as zero-knowledge proofs \cite{fiege1987zero} and differential privacy \cite{wei2020federated}, also hold promise for improving data protection.

In the wider context, the long-term stability of federated military LLMs depends on standard security practices and close cooperation among allied nations.
Future studies should examine how the proposed defenses work in different coalitions, considering policy differences, resource limits, and varying technology levels.
Further research into homomorphic encryption \cite{yi2014homomorphic}, blockchain audits \cite{nofer2017blockchain}, and better anomaly detection methods are needed to strengthen system integrity.
At the same time, adopting global standards for security reviews, model checks, and joint threat response can unify coalition efforts and address evolving prompt injection threats.
Together, these measures will build a stronger and more reliable foundation for federated military LLMs, ensuring their trustworthiness in a contested and complex environment.

\section{Conclusion}\label{sec:conclusion}
In this work, we present potential prompt injection vulnerabilities within federated military LLMs, pinpointing four key attack scenarios.
We highlight the wide-ranging impact of these threats in military settings by presenting how secret data can be leaked, free-riders can exploit the system, operations can be disrupted, and false information can spread.
In response, we introduce a collaborative human–AI framework which combines both technical and policy perspectives, enabling agile defense against rapidly evolving attack patterns.  
Our findings emphasize the need for continual improvement and joint governance to safeguard federated military LLMs against evolving prompt injection attacks.

\bibliographystyle{IEEEtran}
\bibliography{reference}

\end{document}